\documentclass[lettersize,journal]{IEEEtran}
\usepackage{amsmath,amsfonts}
\usepackage{algorithmic}
\usepackage{algorithm}
\usepackage{array}
\usepackage[caption=false,font=normalsize,labelfont=sf,textfont=sf]{subfig}
\usepackage{textcomp}
\usepackage{stfloats}
\usepackage{multirow}
\usepackage{url}
\usepackage{verbatim}
\usepackage{graphicx}
\usepackage{cite}
\begin{document}

\title{TextBlockV2: Towards Precise-Detection-Free Scene Text Spotting with Pre-trained Language Model}
\author{Jiahao Lyu, Jin Wei, Gangyan Zeng, Zeng Li,
Enze Xie,
Wei Wang, 
Yu Zhou
\thanks{
J. Lyu, Z. Li, E. Xie, and Y. Zhou are with the Institute of Information Engineering, Chinese Academy of Sciences, also with the School of Cyber Security, University of Chinese Academy of Sciences, Beijing 100089, China, E-mail: \{lvjiahao, lizeng, xieenze, zhouyu\}@iie.ac.cn. 

J. Wei is with the Lenovo Research, Beijing 100094, China, Email: weijin4@lenovo.com.

G. Zeng is with the School of Cyber Science and Engineering, Nanjing University of Science and Technology, Nanjing 210094, China, Email: gyzeng@njust.edu.cn.

Wei Wang is with Shanghai Artificial Intelligence Laboratory, Shanghai 200043, China, Email: wangwei@pjlab.org.cn.

Y. Zhou is the corresponding author.
}
}

\markboth{JOURNAL OF LATEX CLASS FILES,~Vol.~0, No.~0, March~2024}%
{Shell \MakeLowercase{\textit{et al.}}: TextBlockV2: Towards Precise-Detection-Free Scene Text Spotting with Pre-trained Language Model}


\maketitle

\begin{abstract}
Existing scene text spotters are designed to locate and transcribe texts from images. 
However, it is challenging for a spotter to achieve precise detection and recognition of scene texts simultaneously.  
Inspired by the glimpse-focus spotting pipeline of human beings and impressive performances of Pre-trained Language Models (PLMs) on visual tasks, we ask: 1) ``Can machines spot texts without precise detection just like human beings?", and if yes, 
2)``Is text block another alternative for scene text spotting other than word or character?"
To this end, our proposed scene text spotter leverages advanced PLMs to enhance performance without fine-grained detection.
Specifically, we first use a simple detector for block-level text detection to obtain rough positional information. Then, we finetune a PLM using a large-scale OCR dataset to achieve accurate recognition. 
Benefiting from the comprehensive language knowledge gained during the pre-training phase,
the PLM-based recognition module effectively handles complex scenarios, including multi-line, reversed, occluded, and incomplete-detection texts.
Taking advantage of the fine-tuned language model on scene recognition benchmarks and the paradigm of text block detection, extensive experiments demonstrate the superior performance of our scene text spotter across multiple public benchmarks.
Additionally, we attempt to spot texts directly from an entire scene image to demonstrate the potential of PLMs, even Large Language Models (LLMs).
\end{abstract}


\section{Introduction}
\label{sec:intro}

\IEEEPARstart{S}{cene} text spotting\cite{liu2020abcnet, wang2020ae, zhang2022text, wang2022tpsnet, shu2023perceiving} consists of scene text detection\cite{dai2021accurate, qin2021mask, zhang2022kernel, shu2023ei, qin2023towards, yang2023zoom} and 
scene text recognition\cite{qiao2020seed, qiao2021pimnet, du2022svtr, zhang2023glalt}, and has recently gained significant attention in academia and industry.
Many down-stream tasks regard scene text spotting as an essential step, including table structure recognition\cite{shen2023divide}, visual information extraction\cite{wang2022lilt}, document analysis\cite{da2023vision, yang2023mask},  text-base visual question answering\cite{zeng2023beyond, zeng2023filling}, etc.




\begin{figure}[ht]
\centering
\includegraphics[width=0.5\textwidth]{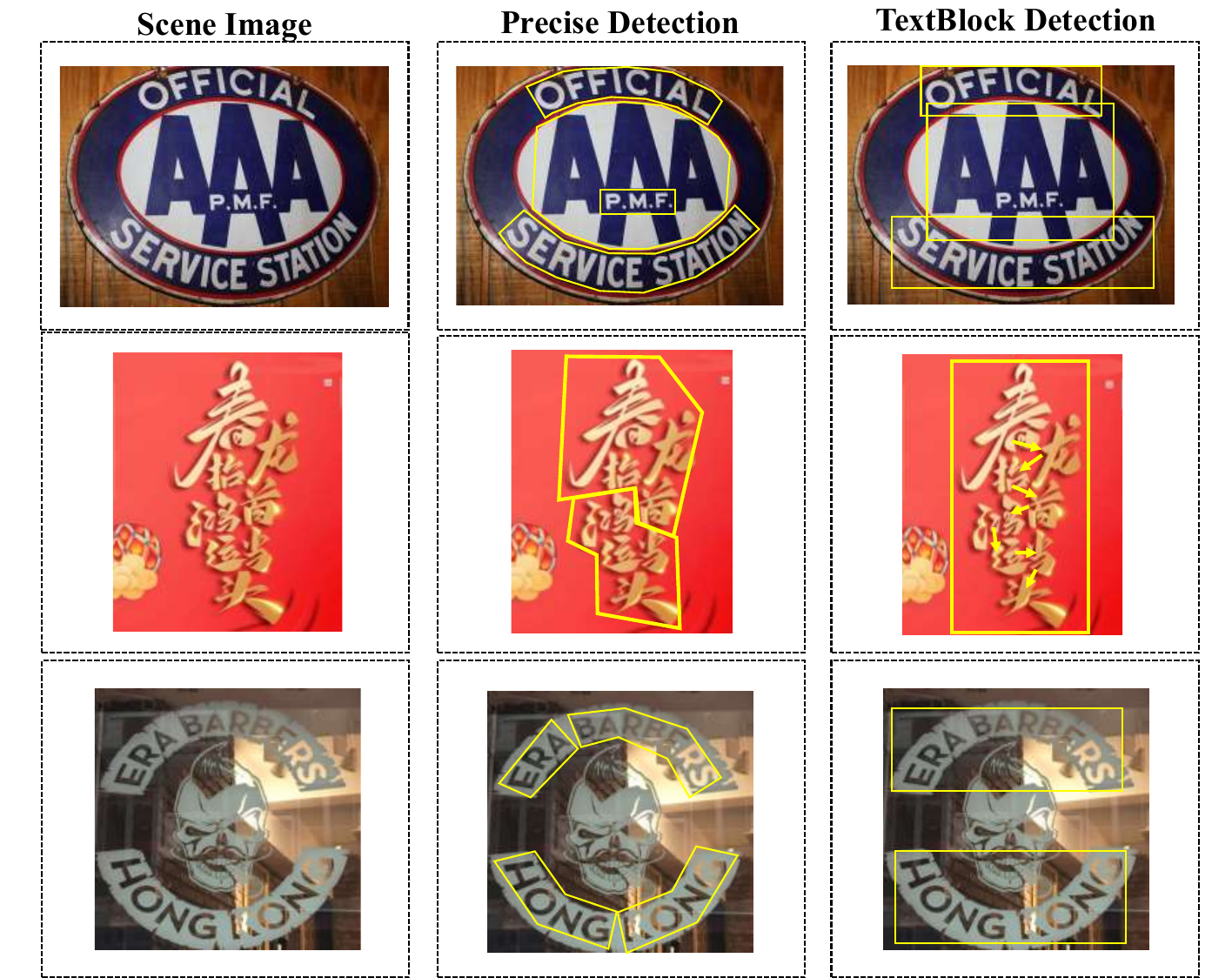}
    \caption{Illustration of Precise Detection and TextBlock Detection. Precise detection aims to detect text units, such as words or phases as shown in the second column. Our proposed detection method, based on text block, reduces the difficulty of detection. The yellow arrows represent the natural reading order in Chinese.}
    \label{fig:1}
\end{figure}

Based on the focus of model design, existing methods can be broadly classified into two branches: detection-oriented scene text spotting and recognition-oriented scene text spotting. Detection-oriented methods follow the detection-first paradigm and aim to obtain accurate text boundaries. For instance,  Mask TextSpotter series \cite{lyu2018mask, liao2019mask,liao2020mask} are derived from general object detection methods,  and several methods\cite{liu2020abcnet,liu2021abcnet,wang2022tpsnet,liu2023pbformer} propose text-oriented representations for the compact text boundaries. 
Nevertheless, recognition-oriented methods \cite{fang2022abinet++, Qin2019TowardsUE, wei2022textblock, qiao2021mango, wang2021implicit}  propose novel and effective recognition methods to alleviate the dependence on precise detection. 



Although existing methods have reported impressive performances, some essential problems remain to be addressed, as shown in Fig. \ref{fig:1}. 
Firstly, obtaining fine-grained precise detection from text instances in formidable layouts can be challenging. The first two images in Fig. \ref{fig:1} underscores the intricacies of precise detection, both existing in English and Chinese. In the first image, the text ``AAA'' and ``P.M.F.'' are coupled together, thus these two instances cannot easily be precisely detected at the same time. The second image encounters the aggregated Chinese characters, which increases the difficulty of precise detection significantly.  
Furthermore, precise detection could lose contextual semantic information because of independent detection results.  This issue is exemplified in the last image of Fig. \ref{fig:1}, where if ``KONG'' is recognized separately, it may be misinterpreted as ``LONG'' due to reflection noises. However, considering ``HONG'' and ``KONG'' together can avoid this error as ``KONG'' is more semantically related to ``HONG'' than ``LONG''.


After considering the previous frameworks and existing problems, we resort to finding a solution from the bionics to solve the above two problems. When reading texts in natural scenes, human beings do not need to accurately locate the specific text instances but only attend to the rough positions and leverage the contexts to read out the contents. Inspired by this phenomenon, we try to throw and answer two questions: \textbf{1) ``Can machines spot text without accurate detection just like human beings?", and if yes, 2) ``Is text block another alternative for scene text spotting other than word or character?".} 

To answer these questions, we propose a novel scene text spotting framework that reduces the reliance on accurate detection. The architecture comprises a block-detection module and a recognition module. 
For block-level detection, we assume that text instances within a block have adjacent positions and similar visual features. By considering both positional and visual features, we cluster visually similar and closely positioned text instances into a single text block. This clustering annotation is then used to train a simple detector without many bells and whistles. The detection module can obtain coarse but high-recall-rate boundaries due to the unconstrained tightness. The novel text block generation algorithm also alleviates the ambiguity of the text block.

During the recognition phase, how to handle various situations in the context of blocks ultimately determines the spotting performance. Factors such as background noises from coarse detection, flexible text arrangements, and long sequence lengths contribute to the difficulties in recognition. To address these challenges, we treat scene text recognition as a sequence modeling task and leverage Pre-trained Language Models (PLMs) to solve it. Without complex design modifications, we fine-tune a PLM using OCR datasets. To tackle the slow convergence problem across domains, we reconsider the relationship between vision and language models and put forward a novel unified vision-language mask (UVLM). This subtle design enhances both recognition performance and convergence speed. With the combination of pre-trained knowledge and elaborate designs, our PLM-powered recognizer achieves higher accuracy in processing complex situations compared to previous methods.

In conclusion, we propose an advanced framework TextBlockV2, the latest version based on our earlier work TextBlock\cite{wei2022textblock} (ACM MM 2022 paper). Considering the room for improvement in the heuristics methods of text block generation and the position-aware recognizer, we further upgrade our method. The differences can be concluded in the following aspects: (1) We notice the problem with the ambiguous definition of text block regions, so we propose a new text block generation scheme, which takes a clustering method that considers the spatial position features and visual features. (2) Compared to the recognition module in TextBlock, we replace it with more powerful PLMs. PLMs introduce prior language knowledge, improving the text recognition task performance. Furthermore, we propose the novel unified vision-language mask (UVLM) to enhance convergence and effectiveness. Extensive experiments show that our modifications of the recognition module are effective. (3) More experiments and analyses are conducted, claiming that our proposed method is effective and superior to the alternative spotting methods.

The main contributions of our work are summarized as four-fold:

\begin{itemize}

\item After rethinking the conventional scene text spotting framework that combines fine-grained detection and isolated-instance recognition, we propose TextBlockV2, the human-like coarse-grained detection and the united multi-instance recognition framework upon the foundation of TextBlock, which relieves the burden of detection and utilizes the rich prior language information to solve difficult situations for recognition simultaneously.


\item To alleviate the ambiguity of the text block region definition, we propose a novel block generation algorithm that leverages the positional relationships between scene texts in the detection phase. By considering both the spatial positions and visual features, we alleviate the burden on the text detector and reduce ambiguity in defining text blocks. This novel algorithm utilizes clustering techniques to improve the precision of block generation.

\item For the recognition module, we fine-tune Pre-trained Language Models (PLMs) to create a powerful recognizer. By leveraging the pre-trained knowledge within PLMs, our recognizer is capable of handling complex situations in text recognition. Additionally, we design a unified vision-language mask (UVLM) for PLM fine-tuning in the scene text recognition task, resulting in faster convergence and improved performance.

\item Our method achieves competitive or even superior performance without relying on accurate text detection, as demonstrated by quantitative experiments on three public benchmarks. Furthermore, we explore the potential of entirely detection-free spotting by experimenting with PLMs, showcasing the versatility and efficacy of our approach.

\end{itemize}

\section{Related Works}

\begin{figure*}[h]
\centering
\includegraphics[width=1.0\linewidth]{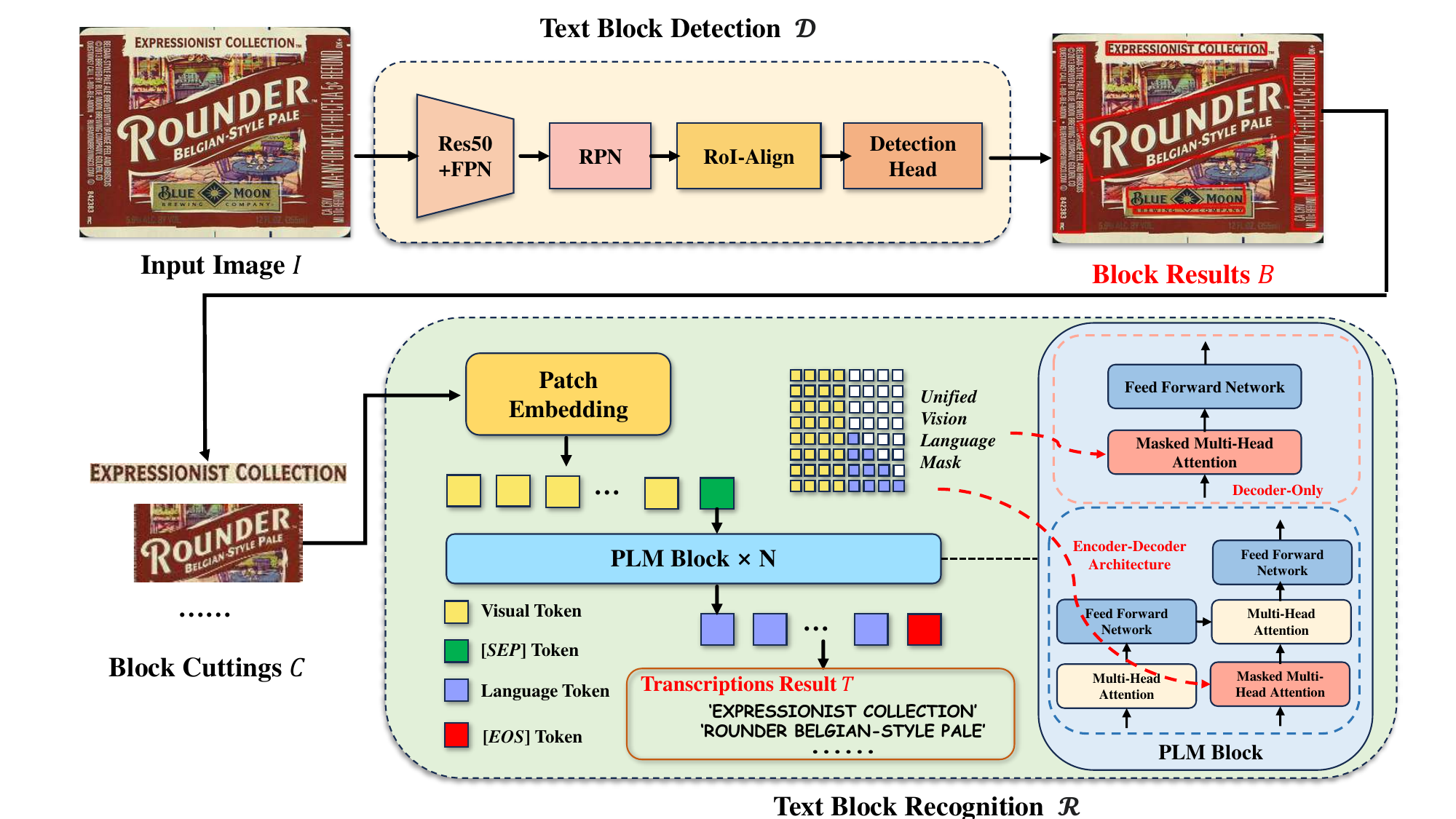}
    \caption{The overview pipeline of TextBlockV2. The scene text image is fed into the TextBlock detection module, which is implemented by Mask R-CNN\cite{he2017mask}. Then block cuttings are patched to visual tokens. The Pre-trained Language Model is regarded as a scene text recognizer, extracting texts from block cuttings. The PLM block can be decoder-only or encoder-decoder architecture.}
    \label{fig:2}
\end{figure*}

\subsection{Scene Text Spotting}
Scene text spotting is a task that involves detecting and recognizing text in natural scenes simultaneously. Existing methods can be classified into two main categories based on their design focus: detection-oriented scene text spotters and recognition-oriented scene text spotters.

\subsubsection{Detection-Oriented Scene Text Spotter}
Detection-oriented Scene Text Spotter follows the detection-then-recognition paradigm. In the preceding stage, the performance of scene text detection plays a vital role in the whole spotting task. Some early detection-oriented spotters derive from general object detection or instance segmentation, such as MaskTextSpotter series\cite{lyu2018mask, liao2019mask, liao2020mask} learned from Mask R-CNN\cite{he2017mask} and TESTR\cite{zhang2022text} drawn on Deformable DETR\cite{zhu2020deformable}. 
In addition, some methods \cite{feng2019textdragon, liu2020abcnet, liu2021abcnet, wang2022tpsnet, liu2023pbformer}  explore a more accurate representation of text boundaries to improve detection performance. These techniques aim to precisely delineate the boundaries of individual characters or words, enhancing the overall quality of text detection. 
Moreover, a few methods\cite{xing2019convolutional, baek2020character, wang2021pan++} adopt bottom-up paradigms for scene text detection. These approaches first predict individual characters or text components and then connect them to form complete text instances. This strategy allows for flexible and efficient detection of text in complex scenes.
Recently, Transformer-based architectures are applied to enhance the detection performance of scene text spotters, such as SwinTextSpotter\cite{huang2022swintextspotter}, SPTS\cite{peng2022spts, liu2023spts}, DeepSOLO\cite{ye2023deepsolo}, ESTextSpotter\cite{huang2023estextspotter}.  These models benefit from the powerful representation learning capabilities of Transformers, enabling them to capture critical textual patterns and context for accurate detection. 
While impressive performance has been achieved by these detection-oriented scene text spotters, precise detection remains a critical bottleneck in the spotting process. Therefore, we attempt to detect
texts coarsely and recognize multiple text instances simultaneously,
which shifts the burden from detection to recognition.

\subsubsection{Recognition-oriented Scene Text Spotter}
In addition to detection-oriented approaches, recognition-oriented scene text spotters also hold a prominent position in the field of text spotting. \cite{li2017towards}  proposes a unified network that performs simultaneous text detection and recognition in a single forward pass. The approach shares convolution features and employs a curriculum strategy based on CRNN\cite{shi2016end}. \cite{he2018end} focuses on explicit alignment and attention to boost the performance of recognition. 
\cite{wang2021implicit} notices the problem that the recognition module is constrained to transcribe from the cropped text images, and proposes the Implicit Feature Alignment to free the detection module in the inference stage. However, character-level annotations and complicated post-processing are necessary. MANGO \cite{qiao2021mango} proposes the position-aware mask attention module to generate the attention weights for each instance and character and allocate different text instances into different feature map channels.  Then a lightweight sequence decoder is used to decode the character sequences. ABINet++\cite{fang2022abinet++} extends from ABINet\cite{fang2021read}, a text recognizer leveraging language information. \cite{tang2022you} uses voice annotation to assist in recognizing scene text better. Motivated by the glimpse-focus spotting mechanism observed in human beings and the impressive capabilities of pre-trained language models, we opt to develop a robust recognizer to elevate the entire scene text spotting process, free from the constraints of precise detection.

\subsection{{Enhancing Vision Tasks with Pre-trained Language Models}}

Pre-trained Language Models have been widely applied in various NLP tasks, following the paradigm of pre-training and fine-tuning for learning. 
As one of the pioneers in PLMs, ELMo\cite{matthew2018peters} is introduced to capture contextual representations through a bidirectional LSTM network. Transformer facilitates the enhancement of various downstream tasks through the utilization of encoder-only\cite{devlin2018bert}, decoder-only\cite{radford2018improving}, and encoder-decoder architectures\cite{raffel2020exploring}. Besides, numerous language-aware vision tasks aim to leverage the linguistic knowledge within PLMs. 
For example, \cite{shen2022k, ma2023borrowing, wang2023actionclip} explore novel approaches that leverage linguistic knowledge for visual learning. DTrOCR\cite{fujitake2023dtrocr} firstly attempts to fine-tune GPT2 to recognize texts in various scenes and languages. FITB\cite{zeng2023filling} formulates the TextVQA task as the ``Filling in the Blank" problem by employing prompt-tuning, which is a widely adopted training strategy in pre-trained language models.

Furthermore, Large Language Models (LLMs) are Transformer-based language models with billions of parameters\cite{zhao2023survey}. Recent studies have shown large language models emerge with some magic abilities that are not presented in small models. Taking into account computational resources and data volume, we undertake the redesign of PLMs as an initial exploration of their potential in OCR tasks.

\section{Methodology}

As shown in Fig. \ref{fig:2}, we propose a two-stage scene text spotting pipeline without precise detection, named TextBlockV2. In this section, we provide a brief overview of the entire architecture. Next, we first introduce a novel approach for block-level label generation and the PLM-based block recognition, enhanced with the Unified Vision-Language Mask. Lastly, we will outline the training and inference procedures in detail.

\subsection{Overall Architecture}
As illustrated in Fig.\ref{fig:2}, the whole architecture comprises a detection module $\mathcal{D}$ and a recognition module $\mathcal{R}$. Without whistles and bells, the Mask R-CNN with ResNet50 and FPN is regarded as the block detector $\mathcal{D}$. Given a scene text image $I$,  $\mathcal{D}$ extracts block-level results $B = \{B_i\}_{i=1}^n$, where $n$ is the number of text blocks in $I$. To train $\mathcal{D}$, which can localize the block-grained texts, we utilize data generated by a novel text block generation algorithm, which is described in \ref{detection-label-generation}.
Subsequently, the block cuttings $C = \{C_i\}$ undergo cutting and rectifying processing before being fed into $\mathcal{R}$.  Implemented by PLM, $\mathcal{R}$ is responsible for recognizing the various words presented in the images. The transcriptions of $\mathcal{R}$ can be represented by $T = \{T_i\}$. We enhance the performance of $\mathcal{R}$ by incorporating the Unified Vision-Language Mask. The spotting results include $B$ and $T$.

For the details of $\mathcal{R}$, the block cuttings $C$ are resized to $H \times W$ and patched into visual tokens $V = \{V_j\}_{j=1}^m$ using a patch embedding module, where $m$ is the length of visual tokens.
The patch width and height are denoted as
$H_p$ and $W_p$ respectively, thus the number of visual tokens can be calculated as $m=\frac{H \times W}{H_p \times W_p}$. 
After that, a dedicated token known as the $[SEP]$ token is appended after the visual tokens to indicate the boundary between vision and language tokens. Subsequently, the language tokens $L = \{L_k\}_{k=1}^{n}$ are generated by the PLM blocks in an auto-regressive manner until the End-Of-Sequence $[EOS]$ token is encountered. The final transcription results $T$ are obtained from the language tokens $L$ using beam search. For the PLM block, we try on two main-stream architectures, which are encoder-decoder and decoder-only architectures. As the vital module of cross-modal interaction, the masked multi-head attention module is necessary for both architectures. With the masked multi-head attention module, we propose a Unified Vision-Language Mask (UVLM) that effectively utilizes both vision and language tokens. Details about UVLM will be discussed in Section \ref{uvlm}.

\subsection{Detection Label Generation}
\label{detection-label-generation}

\begin{figure}[t]
\centering
\includegraphics[width=0.5\textwidth]{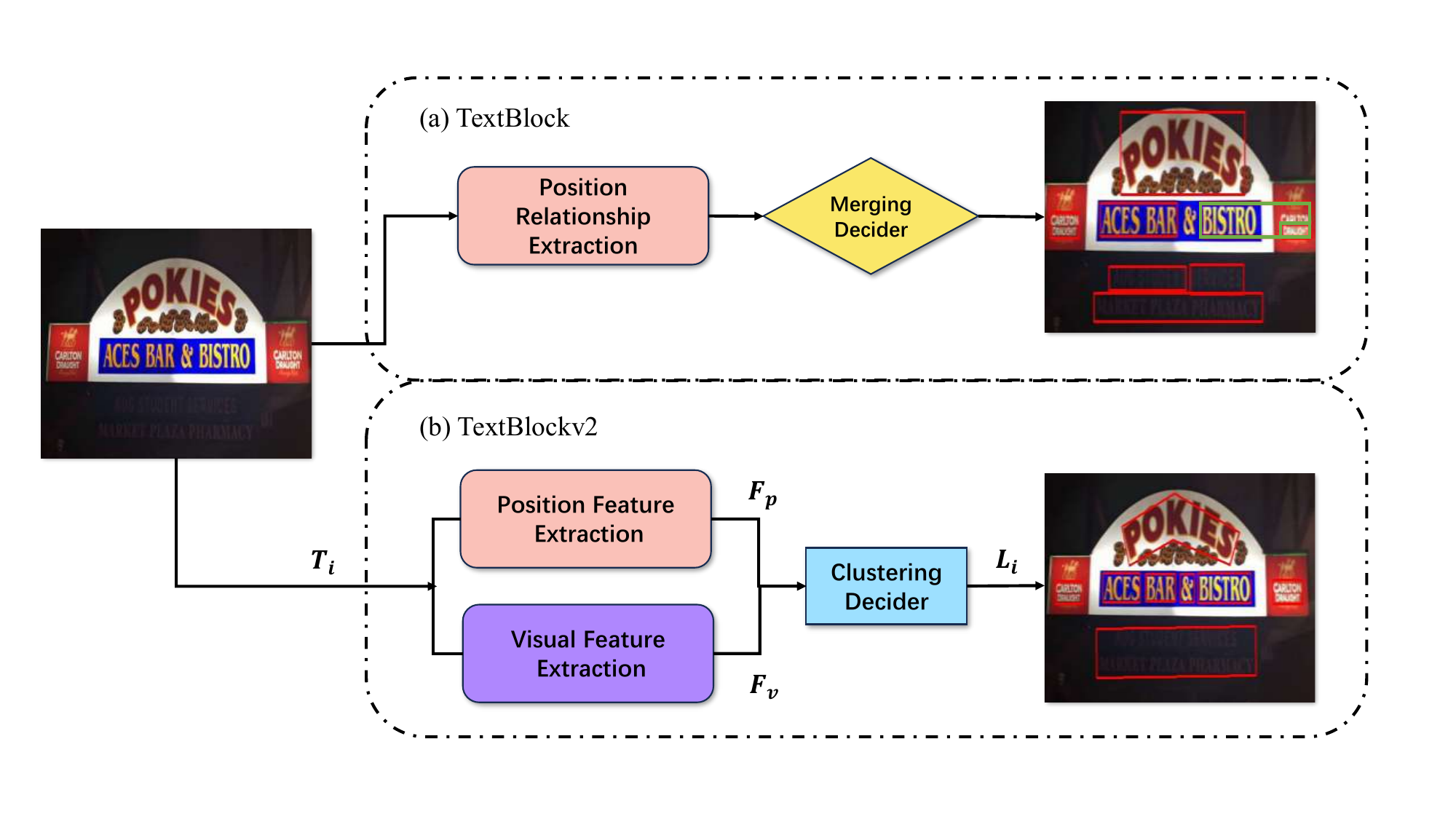}
    \caption{Comparison of the text block generation pipeline between TextBlock and TextBlockV2. The red boxes represent semantically appropriate text blocks, while the green ones do not.}
    \label{fig:3}
\end{figure}

Achieving precise detection results remains challenging for scene text detectors. Therefore, how to define an easy-to-detect block is a crucial problem.
In TextBlock, the text block generation algorithm only relies on the position relationship between text instances, as depicted in Fig. \ref{fig:3}(a). However, this heuristic method often produces ambiguous instances, which in turn poses challenges for the training of text block detectors.  
To mitigate the uncertainty in the block generation phase, we replace the heuristic text block generation method with a clustering algorithm that considers both position features and visual features simultaneously.

Considering the text instances within a block are expected to exhibit similar spatial positions and visual features, we propose a novel text block generation algorithm. Given an image $I \in \mathcal{R}^{h \times w \times c}$ and $N$ text instances represented as $\textbf{T} = \{\textbf{T}_i\}_{i=1}^n$, the coordinates of each text instance are denoted as $\mathcal{C}_j = \{(x_j, y_j)\}_{j=1}^{n}$ where $n$ is the number of polygon vertices. We calculate the position features $F_p \in \mathcal{R}^{n}$ using Eq. \ref{eq:f_p}, which aims to normalize the text coordinates to facilitate further analysis and clustering.

\begin{equation}
   F_p = \{x_j / w, y_j / h\}
\label{eq:f_p}
\end{equation}

In addition to spatial positions, visual features are also important for distinguishing text blocks.  
For each text instance $\textbf{T}_i$, we leverage the pre-trained backbone oCLIP \cite{xue2022language} to extract the text-aware features. The visual features $F_v \in \mathcal{R}^{d}$ can be calculated as Eq. \ref{eq:f_v}, where $d$ represents the dimension of visual features. $GAP(\cdot)$ denotes the global average pooling operation, $oCLIP(\cdot)$ represents feature extraction, and $Preprocess(\cdot)$ encompasses pre-processing operations such as \textit{Resize}, \textit{Normalize} and \textit{To-Tensor}.  By computing the visual features in this manner, we can effectively capture the visual characteristics of each text instance, providing a more reasonable reference for the subsequent clustering process.

\begin{equation}
   F_v = GAP(oCLIP(Preprocess(\textbf{T}_i)))
\label{eq:f_v}
\end{equation}

Furthermore, we concatenate the position feature $F_p$ and visual feature $F_v$ into $F_i \in \mathcal{R}^{n+d}$ for each text feature. Then we cluster all text features $F$ with DBSCAN\cite{ester1996density} and assign cluster labels $L=\{L_i\}_{i=1}^n$. Text instances with the same label are then merged into a new text block. We utilize the minimization convex polygon approach to determine the merging result. The visualization of our text block generation is presented in Fig. \ref{fig:3}(b), where adjacent and visually similar text instances are combined into a new text block.










 
        

 


\subsection{PLM Recognition Block}
\label{uvlm}
Although coarse-grained detection improves the recall of detection results, it introduces background noises for recognition. Therefore, a robust recognizer should be proposed to handle various challenging situations, to ensure the performance of the whole pipeline.

To tackle these challenges, PLM is introduced to recognize text in these various scenes. There are several reasons why we propose the PLMs-based recognizer to solve the challenging scene text recognition task. Firstly, PLMs are composed of Transformer-based blocks and pre-trained on the massive corpus. The pre-trained weight provides rich language prior, which assists the recognizer in generating more semantic transcription. 
Moreover, Transformer-based blocks have long-range dependencies, which are beneficial for learning powerful and robust representations to recognize complicated scenarios, including multi-line, incomplete-detection, and occluded text instances. 

This section explains how PLMs are applied to the scene text recognition task and introduces a detailed Unified Vision-Language Mask that aims to enhance the performance of the recognizer.  In this paper, two types of PLMs are explored: decoder-only architecture and encoder-decoder architecture, as illustrated by the blue area in Fig. \ref{fig:2}. 

\subsubsection{Unified Visual-Language Mask}
In the Transformer decoder, the mask used in the Masked Multi-Head Attention operation plays a crucial role and greatly impacts the prediction performance. The typical Masked Multi-Head Attention can be conducted as Eq. \ref{eq:mmha}, where $Q, K, V$ represent the Query, Key, Value matrix, $d$ is the dimension of the hidden states, and $\mathcal{M}$ is the binary mask.

\begin{equation}
    Attention = \frac{Softmax(QK^T\bigodot\mathcal{M})}{\sqrt{d}}V
\label{eq:mmha}
\end{equation}

For different vision and language tasks, different attention masks are required. Fig. \ref{fig:3} illustrates the representation of typical vision and language masks, denoted as (a) and (b) respectively. 
In the vision modality, bidirectional visual tokens should be visible to capture the continuity of visual scenes. Therefore, the visual mask can be represented as $\mathcal{M}_V = \mathbf{1}^{N \times N}$, where $N$ is the number of tokens. 
However, in the language modality,  a typical causal mask is employed to ensure that each token can only attend to the preceding tokens in the NLP task. The language mask $\mathcal{M}_L$ is structured as a lower triangular matrix, as shown in Eq. \ref{eq:language_mask}, where $l_i$ and $l_j$ are the indexes of input tokens and output tokens in the language modality.

\begin{equation}
    \mathcal{M}_L=\left\{
\begin{array}{rcl}
0 & & {l_i \leq l_j}\\
1 & & {l_i > l_j}\\
\end{array} \right.
\label{eq:language_mask}
\end{equation}

To make full use of the characteristics of both the vision and language modalities, inspired by the prefix mask in NLP \cite{dong2019unified}, we propose a Unified Vision-Language Mask, denoted as $\mathcal{M}_{VL}$. 
In the vision part, we maintain bidirectional attention, while in the language part, we follow the causal mask. Eq. \ref{eq:vl_mask} can represent this mask, where $v_i$ represents the index of input tokens in the visual modal, $v_n$ is the length of vision tokens, and $l_i$, $l_j$ are the index of input tokens and output tokens in the language modal.

\begin{equation}
    \mathcal{M}_{VL}=\left\{
\begin{array}{rcl}
0 & & {l_i > v_n \bigcap l_i < l_j} \\
1 & & {otherwise}\\
\end{array} \right.
\label{eq:vl_mask}
\end{equation}

\begin{figure}[t]
\centering
\includegraphics[width=0.5\textwidth]{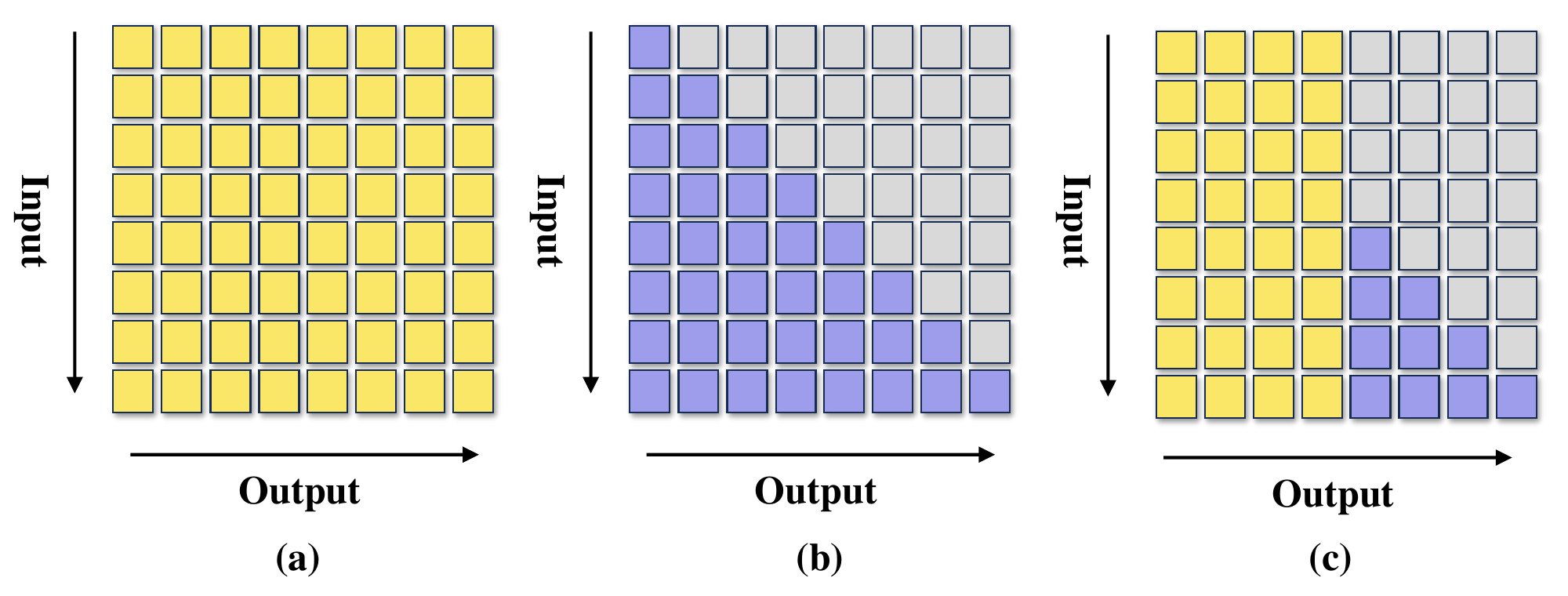}
    \caption{Comparison of three types of masks: yellow, blue, and grey blocks represent vision tokens, language tokens, and masked tokens, respectively. (a) corresponds to the typical visual mask used for bi-directional attention. (b) represents the typical causal mask utilized in language models. (c) signifies our proposed unified vision-language mask that takes into account both vision and language characteristics.}
    \label{fig:4}
\end{figure}

It's worth noting that while the form of UVLM is similar to the prefix mask in NLP, UVLM's goal is to convert text block recognition as an image-to-text translation task which considers both bidirectional attention of vision modality and casual relationship in language modality, as shown in Fig. \ref{fig:4}(c). Additionally, it can be applied to various types of Transformer architectures. Ablation experiments demonstrate the effectiveness of our proposed mask.

\subsubsection{Training and Inference} For the training phase, Pre-trained Language Models (PLMs) have acquired extensive linguistic knowledge from large-scale corpora. However, there is still a gap between different modalities. To bridge this gap, we fine-tune the PLM on scene text recognition datasets. This fine-tuning process helps align tokens from different modalities. We employ the maximum likelihood loss of the language model to optimize the model. Given a sequence of language tokens $X = \{x_{l_1}, . . . , x_{l_n}\}$, the task involves predicting the target tokens $x_{l_i}$ based on the preceding tokens $x_{<l-i}$ in a sequence. The loss function can be expressed as shown in Eq. \ref{eq:loss}. It is important to note that the loss is not related to visual tokens.

\begin{equation}
   \mathcal{L}(X) = \sum_{i=l_1}^{l_n} logP(x_{l_i}|X_{<l_i})
\label{eq:loss}
\end{equation}

During the inference stage, the recognition model conducts scene text recognition by utilizing image tokens along with the beginning token $[SEP]$, which serves as a delimiter between vision tokens and language tokens. Then the recognition block proceeds to predict the target tokens in an auto-regressive manner until it encounters the ending token $[EOS]$. The probabilities of the predicted tokens are calculated using a softmax function over the vocabulary.

\section{Experiments and Results}

\subsection{Datasets}

\noindent\textbf{ICDAR2015}\cite{karatzas2015icdar} consists of 1000 training and 500 testing images, which were captured incidentally. The text instances in these images are multi-oriented and suffer from complicated backgrounds with strong motion blur, perspective, and distortion.

\noindent\textbf{Total-Text}\cite{ch2017total} includes 1255 training and 300 testing focused images. The dataset includes horizontal, multi-oriented and curved text instances, which are annotated with word-level polygons. 

\noindent\textbf{SCUT-CTW1500}\cite{yuliang2017detecting} is a curved text benchmark, which consists of 1000 training and 500 testing images. Text is represented by polygons with 14 points at the text-line level.

\noindent\textbf{Curved Synthetic Dataset 150k} \cite{liu2020abcnet} is a synthetic datasets generated from \cite{gupta2016synthetic}, and it includes nearly 150k images that contains straight and curved texts.

\noindent\textbf{Union14M}\cite{jiang2023revisiting} includes 4M labeled images from 14 previous recognition datasets and several organized testing benchmarks. We use the training part of Union14M, termed Union14M-L, to warm up our recognizer in the training phase of the recognizer. And we conduct the recognition experiments on the testing part. 

\noindent\textbf{SynthTiger-4M}\cite{yim2021synthtiger} contains 4M single-line or multi-line images with texts, which SynthTiger generates. We set the maximum word count to five. This dataset is used during the pre-trained stage of the recognizer.

\noindent\textbf{Real Blocks} is derived from three public benchmarks, ICDAR2015, Total-Text and SCUT-CTW1500. This dataset comprises two essential components: block-level detection and recognition annotations. In the detection part, we transform the word-level or line-level annotations into block-level annotations using the method described in \ref{detection-label-generation}. In the subsequent recognition part, we employ the block-level annotations to execute cutting and rectifying operations on the raw images. These operations facilitate the generation of sub-images, which are compiled as the recognition parts of the dataset. Both the detection and recognition components of the dataset are integral to the fine-tuning phases of detection and recognition.

\begin{table*}[t]

\centering
\setlength{\tabcolsep}{6pt}
\caption{Evaluation on six public recognition benchmarks and three block-level datasets in Real Blocks.}
\begin{tabular}{lcccccccc|ccc}
\hline
\multicolumn{1}{c}{\multirow{3}{*}{Methods}} & \multicolumn{8}{c|}{Word-level}                                                   & \multicolumn{3}{c}{Block-level}   \\ 
\cline{2-12} 
\multicolumn{1}{c}{}                         & IIIT-5k & SVT & \multicolumn{2}{c}{IC13} & \multicolumn{2}{c}{IC15} & SVTP & CUTE & ICDAR2015 & Total-Text & CTW-1500 \\
\multicolumn{1}{c}{}                         & 3000    & 647 & 857        & 1015        & 1811        & 2077       & 645  & 288  &       2072    &     2210       & 2658     \\ \hline

CRNN   \cite{shi2016end}                                   &94.6        &  90.7  &  94.1          & 94.5      &   82.0        &    78.5       &  80.6   &  89.1    &      74.8   &       73.5   &    61.0   \\
TrOCR$_{Base}$   \cite{li2023trocr}                                   &    93.4     &  95.2   &     \textbf{98.4}       &      97.4       &       86.9      &  81.2          &   92.1   &   90.6   &    75.3       &        73.6    &    51.6      \\
SVTR$_{Base}$   \cite{du2022svtr}                                    &    96.0   &   91.5  &       97.1     &      95.7       &       85.2      &      83.7      &   89.9   &   91.7   &      76.3     &    75.2  &    52.7      \\
ABINet  \cite{fang2021read}                                     &     \underline{98.6}    &   97.8  &   98.0         &       \underline{98.0}      &       90.2      &     88.5       &    93.9  &    97.7  &        86.6  &     90.1     &         77.9 \\
ParSeq \cite{bautista2022scene}                                      &     \textbf{99.1}    &  \underline{97.9}   &   \underline{98.3}         &       \textbf{98.4}      &           90.7  &      89.6      &   \underline{95.7}   &    \textbf{98.3}  &      87.8     &        \underline{92.0}   &        \underline{80.6}  \\ \hline
TextBlockV2 (T5)                                           &    96.9     &     95.8       &       97.5      &      97.1       &   \underline{92.4}         &   \underline{90.5}   &   94.1   &     91.0 &   \underline{88.9}   &     85.8       &     75.2     \\
TextBlockV2 (GPT2)                                          &    98.0    &   \textbf{98.1}  &      98.0      &      97.7     &       \textbf{93.7}      &      \textbf{92.0}     &    \textbf{96.4}  &   \underline{97.9}   &     \textbf{91.6}     &   \textbf{92.1}       & \textbf{83.1} \\

\hline
\end{tabular}

\label{tab:recog}
\end{table*}

\subsection{Implementation Details}

\noindent\textbf{Detection Module.} The detection model is implemented by Mask R-CNN in MMOCR\cite{kuang2021mmocr}. During the pre-training phase, we use Curved Synthetic Dataset 150k and several public datasets from real scenes to pre-train our detection model. Model training is performed for two epochs with a batch size of 8. We use the Adam optimizer with a learning rate of 1e-3. After pre-training, we fine-tune our detection model using the corresponding detection dataset in Real Blocks for 300 epochs. The optimizer is SGD and the learning rate is 1e-4. During training, the size of the input image is resized to 640, and random cropping and rotating are employed for data augmentation. During inference, for Total-Text and SCUT-CTW1500 datasets, test images have a maximum size of 1080$\times$720. For the ICDAR2015 dataset, the maximum size is 1920$\times$1080.

\noindent\textbf{Recognition Module.} The recognition model in this work is re-implemented using GPT2-Base and T5-Base from the huggingface transformers repository. The training phase is divided into three stages to achieve faster convergence. During the warming-up stage, we only use Union14M-L for our training data. As for the pre-training stage, we add SynthTiger-4M into data in the warming-up stage to pretrain the model. The recognizer is then fine-tuned using the recognition subset of RealBlock. During training, images are resized unconditionally to $64 \times 256$ pixels and the patch size of $8 \times 8$ is used for training.  The AdamW optimizer is employed with a learning rate of 1e-4 in the warm-up and pre-training stage, and the learning rate is reduced to 1e-5 in the finetuning stage. To optimize memory usage, all experiments for the recognition module are trained using mixed precision. which aims to improve the performance and convergence speed of the recognition model.
All experiments are implemented on NVIDIA RTX 3090.

\begin{table*}[t]
\centering
\setlength{\tabcolsep}{6pt}
\caption{Evaluation on benchmarks of Union 14M \cite{jiang2023revisiting}.}
\begin{tabular}{lcccccccc}
\hline
Methods   & Curve & Multi-oriented & Artistic & Contextless & Salient & Multi-Words & General & Avg  \\ \hline
CRNN\cite{shi2016end}      & 19.4  & 4.5            & 34.2     & 44.0        & 16.7    & 35.7        & 60.4    & 30.7 \\
SAR\cite{li2019show}       & 68.9  & 56.9           & 60.6     & 73.3        & 60.1    & 74.6        & 76.0    & 67.2 \\
SATRN\cite{wang2020decoupled}     & 74.8  & 64.7           & \underline{67.1}     & 76.1        & 72.2    & 74.1        & 75.8    & 72.1 \\
SRN\cite{yu2020towards}       & 49.7  & 20.0           & 50.7     & 61.0        & 43.9    & 51.5        & 62.7    & 48.5 \\
ABINet\cite{fang2021read}    & 75.0  & 61.5           & 65.3     & 71.1        & 72.9    & 59.1        & 79.4    & 69.2 \\
VisionLAN\cite{wang2021two} & 70.7  & 57.2           & 56.7     & 63.8        & 67.6    & 47.3        & 74.2    & 62.5 \\
SVTR\cite{du2022svtr}      & 72.4  & 68.2           & 54.1     & 68.0        & 71.4    & 67.7        & 77.0    & 68.4 \\
MATRN\cite{na2022multi}     & \underline{80.5}  & 64.7           & 71.1     & 74.8        & \underline{79.4}    & 67.6        & 77.9    & \underline{74.6} \\
MAERec-S\cite{jiang2023revisiting}  & 75.4  & 66.5           & 66.0     & \underline{76.1}        & 72.6    & 77.0        & 80.8    & 73.5 \\
MAERec-B\cite{jiang2023revisiting}  & 76.5  & 67.5           & 65.7     & 75.5        & 74.6    & \underline{77.7}        & \textbf{81.8}    & 74.2 \\ \hline
TextBlockV2 (T5)      & 73.1  & \underline{72.7}           & 60.8     & 64.4        & 75.6    & 77.0        &  78.0   & 71.7 \\
TextBlockV2 (GPT2)     & \textbf{87.6}  & \textbf{82.0}           & \textbf{72.3}     & \textbf{77.4}        & \textbf{82.1}    & \textbf{86.0}        & \underline{81.2}    & \textbf{81.2}\\ \hline
\label{tab:recunion}
\end{tabular}
\end{table*}

\begin{figure}[t]
\centering
\includegraphics[width=0.5\textwidth]{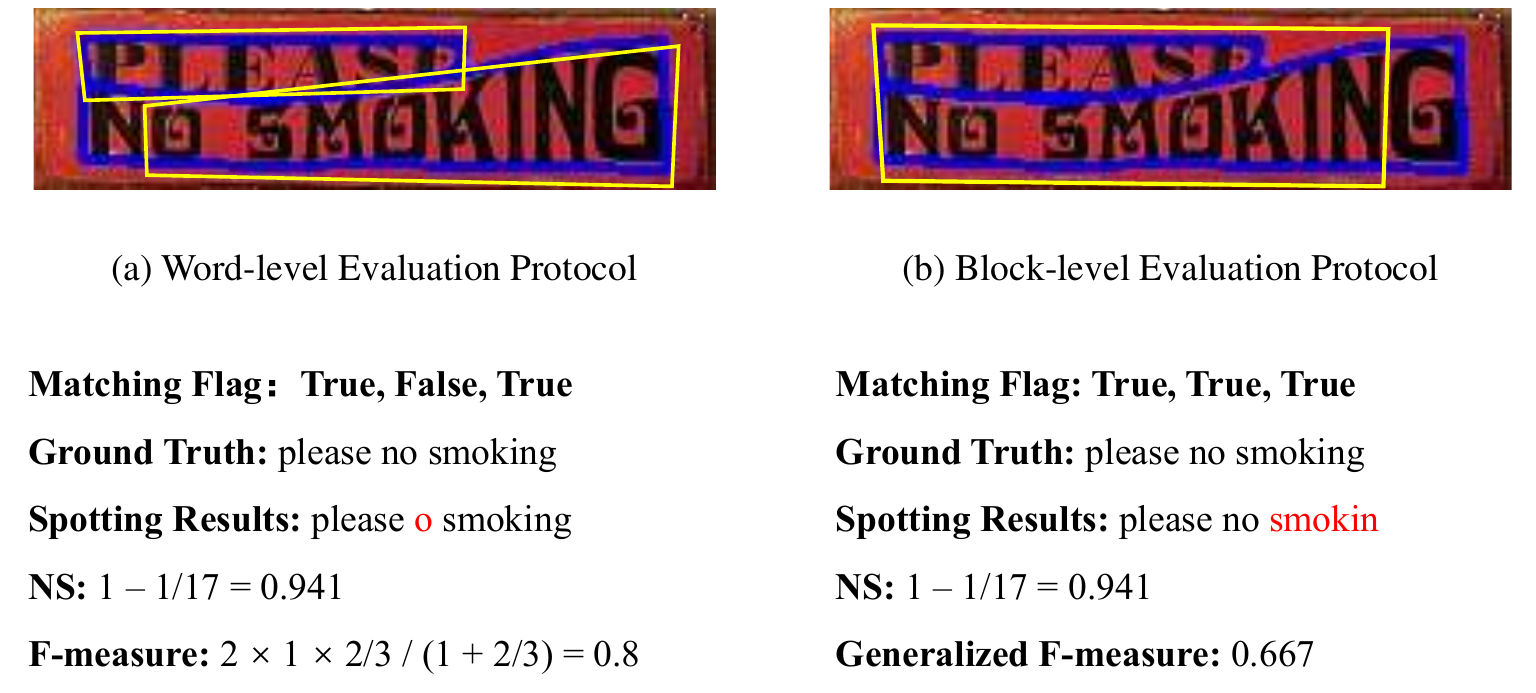}
    \caption{The comparison of two types of evaluation protocols. The blue boundaries are ground truth, and the yellow ones are prediction bounding boxes. This figure shows the detailed calculation of Normalized Scores (NS) and Generalized F-measure (GF). The red texts in the spotting results are incorrect.}
    \label{fig:5}
\end{figure}

\subsection{Evaluation Protocols}
There is an evident difference between block-level and instance-level text spotting. To ensure a fair comparison with existing approaches, two protocols Normalized Score (NS) and Generalized F-measure (GF) are developed.

\noindent\textbf{Normalized Score.}
NS is derived from EEM\cite{hao2021eem}, and advantageous in handling one-to-many and many-to-one matching cases. Specifically, for each image $I$, all ground truth boxes are termed as $G = \{g_j\}_{j=1}^m$, and all prediction boxes are $P = \{p_k\}_{k=1}^n$. The matching process involves two stages. Firstly, a pair matching algorithm is used to match the nearest pair $M'$ for each $g_j$ and $p_k$. Secondly, a set merging algorithm is applied to merge the $M'$ into $M=\{m_i\}=\{(g_i, p_i)\}$ for NS calculation. Suppose that $N$ is the number of $M$ in all test set, the NS can be calculated as Eq. \ref{eq:ns}. Here, $ED(\cdot)$ represents the Edit Distance function, indicating that the NS protocol focuses more on character-level evaluation.

\begin{equation}
    NS = 1 - \frac{\sum_{i=1}^{N}ED(g_i, p_i)}{\sum_{i=1}^{N}max(len(g_i), len(p_i))}
    \label{eq:ns} 
\end{equation}

Note that since the NS score depends on the labels and predictions on one particular dataset, it has no comparability across different datasets.

\noindent\textbf{Generalized F-measure.}
Considering F-measure may not be applicable to the block-level
framework, we propose a novel Generalized F-measure (GF) for fair evaluation. In detail, in the case where a word is correctly recognized, we consider it to be accurately spotted. The matching criterion relies on the geometric relationship between the predicted and ground truth boxes, as defined by Eq. \ref{eq:gf}.

\begin{equation}
    max(\frac{Inter(\alpha(g_i), \alpha(p_j)}{\alpha(g_i)}, 
    \frac{Inter(\alpha(g_i), \alpha(p_j)}{\alpha(p_i)}) > T
    \label{eq:gf}
\end{equation}
where $\alpha(\cdot)$ represents the area of the text instance, and $Inter(\cdot)$ denotes the intersection area of two polygons. The threshold $T$ is set to $0.4$, as referred to TextBlock. 

Further details regarding the evaluation protocols can be found in Fig. \ref{fig:5}. According to Eq. \ref{eq:gf}, the matching scores between each of the 3 ground truth boxes and the yellow box are above the $thr$, so all 3 instances are considered
to be detected correctly, and 2 of 3 are accurately recognized. So the GF is 0.667. It should be noted that GF imposes stricter criteria than the traditional F-measure when assessing the performance of scene text spotting.

\subsection{Block Recognition}
In this section, we evaluate the performance of our PLM-powered recognizer on word-level and block-level datasets, as well as the challenging Union-14M benchmark\cite{jiang2023revisiting}. We use word accuracy as the evaluation metric for all experiments.

Firstly, we compare the encoder-decoder architecture represented by T5 with the decoder-only architecture represented by GPT2. We find that the decoder-only architecture fits the vision-language task better than T5 on all datasets, as shown in Tab. \ref{tab:recog} and Tab. \ref{tab:recunion}. Additionally, Fig. \ref{fig:6} shows that GPT2 is easier to converge in the training phase than T5. Therefore, the default setting of the recognizer is GPT2 in the following experiments if there is no special statement.

Word-level datasets are commonly used as benchmarks in scene text recognition, while block-level datasets are generated from spotting datasets using the method described in Section \ref{detection-label-generation}. The block-level datasets include both single-word images and multi-word images. The recognition results are shown in Tab. \ref{tab:recog} for both word-level and block-level datasets. As can be seen, PLM-based recognizers perform competitively with other methods on word-level benchmarks. In particular, they outperform the state-of-the-art method\cite{bautista2022scene} on ICDAR2015 dataset by an average improvement of $3\%$.

Moreover, we find that our recognizer shows a striking boost for block-level datasets. Specifically, compared with the previous superior method\cite{bautista2022scene}, our recognizer achieves an improvement of $3.8 \%$ on IC15, $0.1 \%$ on Total-Text, and $2.5 \%$ on CTW-1500. Since texts in ICDAR2015 are more incidental and complex, the results suggest that our method can handle more complex situations like multi-word images.

We also conduct another experiment on the latest benchmarks Union14M\cite{jiang2023revisiting} to validate the effectiveness and robustness of our recognizer. Union14M introduces seven subsets to evaluate the robustness of recognition in various situations. Tab. \ref{tab:recunion} shows the experimental results on difficult scenes, and we claim that our recognizer achieves state-of-the-art performances on six out of seven benchmarks. In the difficult subsets, e.g. Curve, Multi-oriented, Multi-Words, our method significantly improves by around 10\% compared to previous methods and exhibits competitive performance on the general subset. The success of our recognizer is attributed to sufficient prior language knowledge of PLM.

\begin{table*}[t]
\centering
\setlength{\tabcolsep}{6pt}
\caption{Scene text spotting results on three public benchmarks. $^\star$ means using Generalized F-measure for evaluation. $^\dag$ means evaluation with $IoU > 0.1$. F-Measure of previous methods is evaluated on the ICDAR2015 dataset with "Generic" lexicons,  Total-Text and SCUT-CTW1500 with "None" lexicon.}
\begin{tabular}{lccccccc}
\hline
{\multirow{2}{*}{Methods}} &  \multirow{2}{*}{Published} & \multicolumn{2}{c}{ICDAR2015}                          & \multicolumn{2}{c}{TotalText}                          & \multicolumn{2}{c}{SCUT-CTW1500}                            \\
& &\multicolumn{1}{c}{NS ($\%$)} & \multicolumn{1}{c}{F-Measure ($\%$)} & \multicolumn{1}{c}{NS ($\%$)} & \multicolumn{1}{c}{F-Measure ($\%$)} & \multicolumn{1}{c}{NS ($\%$)} & \multicolumn{1}{c}{F-Measure ($\%$)} \\ \hline
{Mask TextSpotter v1\cite{lyu2018mask}}      &  ECCV'18   &       58.3            &              62.4                 &            -            &             52.9                  &            -            &     -                          \\
{CharNet Hourglass-88 \cite{xing2019convolutional}}   &  ICCV'19 &            74.2            &        69.1                       &         -               &             -                  &          -              &            -                   \\
{Mask TextSpotter v2\cite{liao2019mask}}  &  TPAMI'19  &         78.1               &             73.5                              &             74.5           &              65.3                 &        -       &     -                      \\
{Mask TextSpotter v3\cite{liao2020mask}} &  ECCV'20   &         78.0               &               74.2                &                         75.3          &              71.2            &       -          &         -                       \\
{ABCNet v2\cite{liu2021abcnet}}      &    TPAMI'21      &          -              &               73.0                &            -            &                 70.4              &               76.4         &              57.5                 \\
{MANGO\cite{qiao2021mango}} $^\dag$    &    AAAI'21           &         75.6               &         69.4                                &           \textbf{82.7}             &               72.9                &            -             &         -           \\
{SPTS\cite{peng2022spts}}       &    MM'22     &           -          &           65.8                   &            -           &             74.2                  &         -              &         63.6\\
{TESTR\cite{zhang2022text}}  &  CVPR'22  &    78.7      &         73.6               &             77.9                  &         73.3               &               79.5                &           56.0               \\
{TextBlock\cite{wei2022textblock}}   &  MM'22           &       79.3                 &            73.7                   &           78.9             &             70.2                  &          77.7              &         56.7\\

DeepSolo\cite{ye2023deepsolo}      &    CVPR'23       &       81.3                 &           79.1                    &           82.6            &                   \textbf{82.5}            &          81.0              &        64.2                       \\
{ESTextSpotter\cite{huang2023estextspotter}}   &  ICCV'23       &      80.1                  &            78.1                   &         81.7              &          80.8                     &            82.9            &                  64.9             \\ \hline
{TextBlockV2(Ours)}                &       -       &      \textbf{82.3}                 &      {\textbf{ 80.5}$^\star$}                    &             82.2           &    { 80.0$^\star$}                          &          \textbf{83.3}             &       \textbf{ 65.1}$^\star$                        \\ \hline
\end{tabular}
\label{tab:sota}
\end{table*}

\subsection{Performances on Scene Text Spotting}
To validate the effectiveness of our block-level pipeline, we compare our method with previous end-to-end scene text spotters on three public benchmarks. The results of the scene text spotting task are shown in Tab. \ref{tab:sota}.

\noindent\textbf{Results on ICDAR2015.}
We conduct experiments on ICDAR2015 to verify the effectiveness of TextBlockV2 on multi-oriented and incidental scene texts. The results show that our method outperforms all previous methods in both NS and F-measure. Specifically, TextBlockV2 achieves the NS of 82.3\% and the GF of 80.5\%, surpassing DeepSolo by 1.0\% and 1.4\%. The reason is that the texts in the ICDAR2015 dataset are small and clustered, making it challenging for word-level pipelines to distinguish between ambiguous samples. In contrast, block units are more reasonable and effective. On the one hand, the ambiguity-less text blocks improve the precision of text block detection. On the other hand, our block-based recognizer learns the semantic information among nearby instances better while previous word-level recognizers focus on isolated-instance prediction.

\noindent\textbf{Results on Total-Text.}
For the Total-Text dataset, scene texts are arbitrarily shaped and clear. Compared with existing end-to-end spotters, our TextBlockV2 achieves competitive performances on NS of 82.2\% and F-measure of 80.0\%. However, our method does not exhibit significant improvement because the scene texts in Total-Text are certain and easy to recognize, and certain text instances do not require merging into a text block. Nevertheless, TextBlockV2 performs considerably better than its previous version, aided by the use of PLM and our novel block generation algorithm.

\noindent\textbf{Results on SCUT-CTW1500.}
Similar to Total-Text, SCUT-CTW1500 comprises numerous arbitrarily shaped and clear texts. However, the use of line-level annotations introduces unique challenges, particularly during the recognition phase. The experiment results indicate that our method outperforms previous approaches in terms of NS and F-measure. From our perspective, the robust performance of our recognizer in managing multi-word instances is the key factor contributing to the success of our spotter on SCUT-CTW1500.

\subsection{Ablations}
In this subsection, we conduct ablation experiments on ICDAR2015 to assess the influence of various components in TextBlockV2. The impact of different core modules is presented in Tab. \ref{tab:ablation}, while Tab. \ref{tab:size} and \ref{tab:training} illustrate the variations in detailed settings.

\begin{table}[t]
\centering
\caption{Ablations on SCUT-CTW1500. The baseline is TextBlock. GPT2 means the recognizer is trained on GPT2 without pre-trained weight. PW means the recognizer is trained with pre-trained weight on the language prediction task. UVLM means the Unified Vision-language Mask. BG means training detector on the annotations from the improved block generation algorithm.}
\begin{tabular}{ccccccc}
\hline
Settings  & GPT2 & PW & UVLM &  BG                   &    NS (\%) & GF (\%) \\ \hline
Baseline &&&&& 77.7  & 56.7   \\
\#1        & \checkmark  &&&            &  78.4   & 57.0  \\
\#2  & \checkmark  & \checkmark  &&  & 81.1  &  61.9      \\ 
\#3     & \checkmark & \checkmark  &\checkmark &  & 81.8   &  63.0     \\ 
\#4     & \checkmark  & \checkmark & \checkmark & \checkmark        &  83.3   & 65.1   \\
\hline
\end{tabular}

\label{tab:ablation}
\end{table}

\begin{table}[t]
\centering
\caption{Comprison with different patch size settings, evaluated on SCUT-CTW1500 subset of Real Blocks.}
\begin{tabular}{c c c | c c}
\hline
Settings & Patch Size    &  Input Size        &   Acc (\%) & FPS \\ \hline

\#1 & $8 \times 8$  &    $64 \times 256$           &    83.1   &  22.3  \\
\#2 & $8 \times 8$   &    $32 \times 256$     &   82.2  &  28.4  \\
\#3 & $8 \times 8$   &    $32 \times 128$    &   80.8  &  34.1  \\
\#4 & $8 \times 4$   &    $64 \times 256$     &   83.3  &  15.0  \\
\#5 & $16 \times 8$   &   $64 \times 256$    &   81.6   &  37.6  \\
 \hline
\end{tabular}

\label{tab:size}
\end{table}

\begin{table}[t]
\centering
\caption{Comprison with different training settings, evaluated on the SCUT-CTW1500.}
\begin{tabular}{l c c | c c}
\hline
Warm-up   &  Pre-training   &   Fine-tuning     &    NS (\%) & GF (\%) \\ \hline
    \checkmark  &   \checkmark   &   &  78.0    &   56.7 \\
 \checkmark  &        &    \checkmark      &  80.5   & 62.7   \\
   &    \checkmark    &   \checkmark &  76.4    &  55.0  \\

\checkmark    &    \checkmark    &   \checkmark &  83.3    &  65.1  \\
\hline
\end{tabular}

\label{tab:training}
\end{table}

\noindent\textbf{Recognizer with PLM.} 
Firstly, we explore the effectiveness of the integration of PLM. As shown in \#1 and \#2 of Tab. \ref{tab:ablation}, even without pre-training weights, GPT2 outperforms the baseline by a small margin, demonstrating that PLM has the potential for scene text recognition. Furthermore, when trained with pre-trained weights, GPT2 exhibits powerful performance, leading to an overall architecture performance of 81.1\% in NS and 61.9\% in GF. Based on these experimental results, we argue that PLM is well-suited as a recognizer after being trained on OCR data.

\begin{figure}[t]
\centering
\includegraphics[width=0.5\textwidth]{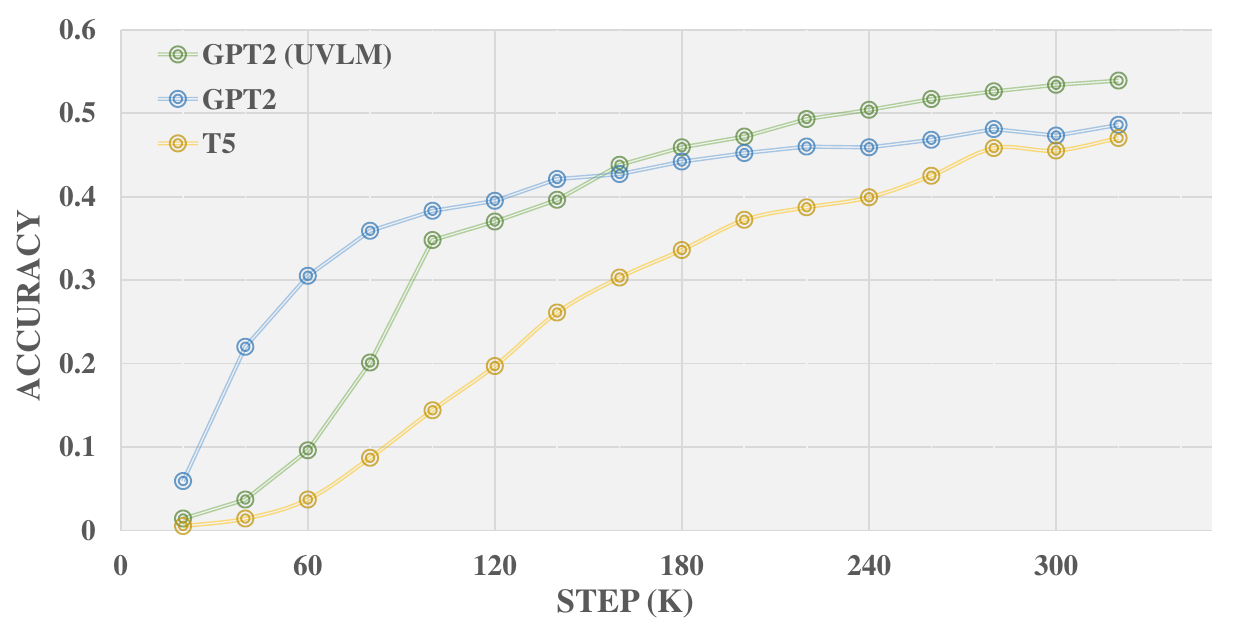}
    \caption{Ablation for the recognition block and Unified Vision-Language Mask (UVLM) on convergence. The Accuracy is evaluated on the recognition task of Real Blocks.}
    \label{fig:6}
\end{figure}

\noindent\textbf{Unified vision language mask.}
The advancement of our unified vision-language mask (UVLM) can be claimed from two perspectives. Firstly, the convergence of our recognizer with the elaborate mask surpasses that of the primary mask used in GPT2. Fig. \ref{fig:6} illustrates the different convergence characteristics when using the UVLM. After incorporating the unified vision-language mask, our modified GPT2 achieves significantly higher recognition accuracy on the testing set compared to the baseline throughout the warming-up stage. This indicates that UVLM makes convergence faster. Secondly, the design of UVLM is specifically tailored for scene text recognition, which serves as a bridge between vision and language modalities. The recognition results from \#3 of Tab. \ref{tab:ablation} demonstrate the superiority of our Unified Vision-Language Mask. Additionally, for the spotting task, UVLM improves NS by 0.7\% and GF by 1.1\%. These experiments validate the efficacy of our elaborate mask for the scene text recognition task, both in terms of convergence and performance.

\begin{figure*}[t]
\centering
\includegraphics[width=1.0\linewidth]{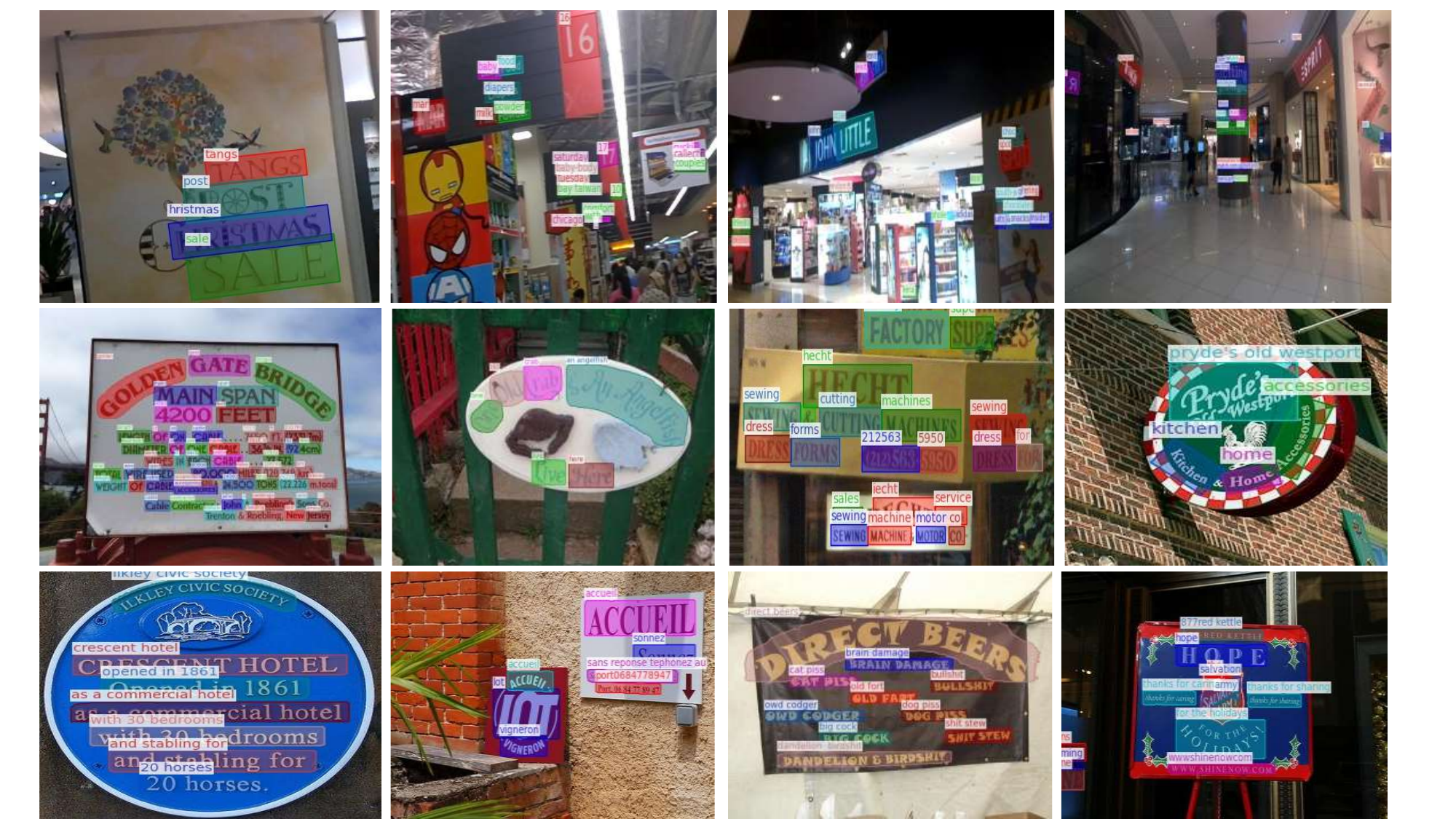 }
    \caption{Visualization results of our TextBlockV2 on ICDAR2015, Total-Text, and SCUT-CTW1500 from top to bottom. Zoom in and out for the best view. }
    \label{fig:7}
\end{figure*}

\begin{figure*}[ht]
\centering
\includegraphics[width=1.0\linewidth]{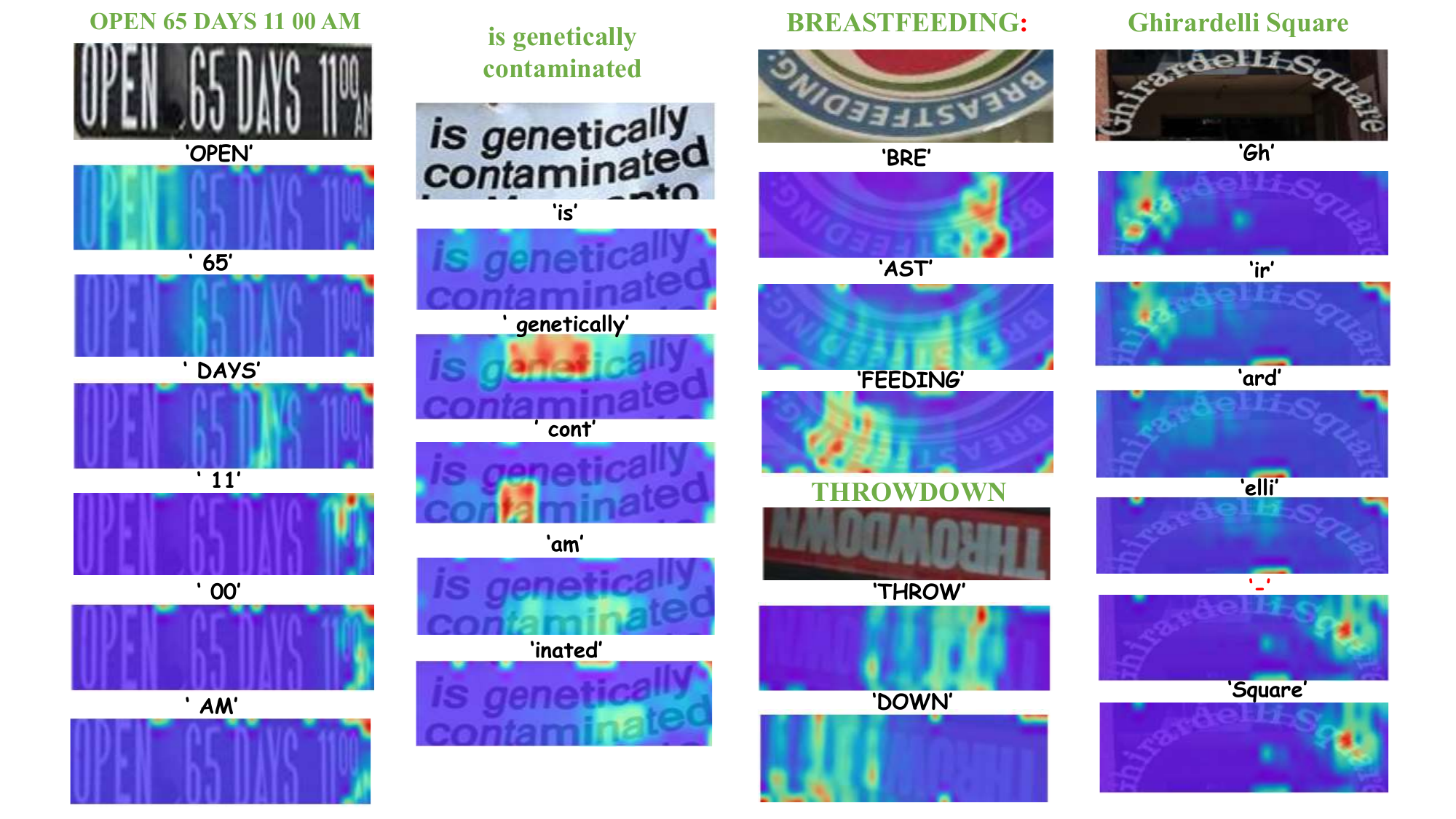 }
    \caption{Visualization of the attention maps during the decoding stage. The attention maps are obtained from the output of the last decoder layer. The green texts are ground truth, the black ones are correct predictions and the red ones are wrong predictions}
    \label{fig:8}
\end{figure*}

\noindent\textbf{Improved block generation algorithm.}
Building upon our previous advancements, we further investigated the positive effect of the improved text block generation algorithm. As indicated in the last line of Tab. \ref{tab:ablation}, incorporating the improved block annotations resulted in an improvement of 1.5\% in NS and 1.9\% in GF. This confirms that our improved block annotations contribute to the enhanced precision of text detection.


\noindent\textbf{The settings of patch size and input size.}
To explore the impact of detailed settings on patch size and input size, we conduct experiments. Firstly, we observe that a larger input size leads to higher performance but slower inference speed, as depicted in \#1, \#2, \#3 of Tab. \ref{tab:size}. Increasing the input size improves overall performance but comes at the cost of increased computational overhead during inference. Additionally, we conduct experiments with patch sizes of $8 \times 4$ and $16 \times 8$. The results of  \#4 and \#5 from Tab. \ref{tab:size} indicate that a smaller patch size can enhance recognition accuracy. However, it also introduces a larger number of vision embeddings, resulting in heavier computational loads during inference. Considering the balance between effectiveness and efficiency, we determine the first line of Tab. \ref{tab:size} as the optimal setting.

\noindent\textbf{The settings of the training phase.} Here, we employ different training strategies to evaluate the effectiveness of each training phase: warm-up, pre-training, and fine-tuning. While a single phase alone may not be meaningful for the overall training process, we explore the impact of combining the training phases in pairs. As shown in Tab. \ref{tab:training}, we find the warm-up and fine-tuning significantly affect the final spotting results, because the model cannot converge entirely without the warm-up phase, and the fine-tuning phase can eliminate the domain gap between the training and testing data. Regarding the pre-training stage, its presence or absence leads to a performance gap of 2.8\% in NS and 2.4\% in GF. This indicates that the inclusion of massive synthetic data in the pre-training stage aids in improving the ability of our recognizer.

\subsection{Qualitative Analysis}

As shown in Fig. \ref{fig:7}, we visualize some spotting examples from three benchmarks. It can
be observed that our method can accurately detect and recognize the texts. Some adjacent and visually similar can be detected in a text block, as shown in the last column of Fig, \ref{fig:7}. Meanwhile, Fig. \ref{fig:8} illustrates the attention maps during the decoding stage. The visualization results indicate that the recognizer accurately focuses on the correct position of sub-words at each step in the reading order. This suggests that using PLM can bridge the gap between vision and language modalities and effectively address scene text recognition task.

\begin{table}[t]
\centering
\caption{The direct spotting results without detection. GPT4-V means the gpt-4-vision-preview model proposed by OPENAI. The prompt is set as "Only Read all texts from the image, Do not output other words." The evaluation method is Word Spotting.}
\begin{tabular}{lccc}
\hline
Methods          & ICDAR2015            & Total-Text & CTW-1500 \\ \hline
NPTS  \cite{peng2022spts}    &   69.4                & 64.7  &  55.4  \\
GPT4-V \cite{openai2023gpt}    &    41.5     &  58.4 & 65.3 \\ \hline
Ours     & 39.6 & 79.9 & 68.9 \\
\hline
\end{tabular}
\label{tab:spot}
\end{table}

\subsection{Detection-Free Spotting}
Furthermore, we attempt to spot using fine-tuned PLM architecture directly without the detection module. The evaluation process adopts GF without the limitation of Eq. \ref{eq:gf}. We regard the NPTS and GPT4-V as competitors, which both spot texts without detection. Tab. \ref{tab:spot} shows the results of detection-free spotting on three public benchmarks. Our method achieves the state-of-the-art on Total-Text and SCUT-CTW1500. As the limitation of the vision token maximum length, our spotter still has a long way to go to the practical detection-free spotter. However, the experiment shows that the PLM recognition module has the preliminary ability to spot scene texts directly. However, how to improve the accuracy of spotting in low-quality situations and save the calculation resources are the potential problems in the next step.

\section{Conclusion} 
In this paper, we propose the TextBlockV2, spotting scene texts without relying on precise detection. To the best of our knowledge, it is the first time to leverage the PLM to the scene text spotting task. Specifically, we propose a novel text block generation algorithm to generate unambiguous annotations, addressing the challenge of ambiguity in text block annotations. Furthermore, we present a unified vision-language mask that enhances the modeling of the relationship between the visual and textual modalities, leading to improved performance. Our experimental results demonstrate that our proposed spotter achieves competitive performance on three public benchmarks. Moreover, we explore the possibility of detection-free spotters using PLMs and even LLMs. In the era of large language models, we aim to further investigate and develop detection-free spotting methods from a practical perspective.

\section*{Acknowledgments}
Supported by National Natural Science Foundation of China (Grant NO. 62376266) and Key Research Program of Frontier Sciences, CAS (Grant NO. ZDBS-LY-7024).



\bibliographystyle{IEEEtran}
\bibliography{ref}



 





\end{document}